\documentclass[letterpaper, 10 pt, conference]{ieeeconf}
\usepackage{algorithm}
\usepackage{algorithmic}
\usepackage{multirow}
\usepackage{hhline}
\usepackage{makecell}
\usepackage{amssymb}
\usepackage{CJKutf8}
\usepackage{adjustbox}
\usepackage{amsmath}
\usepackage{wrapfig}
\usepackage{color}
\usepackage{verbatim}
\IEEEoverridecommandlockouts                              
\title{\LARGE \bf
Fine-Grained Pillar Feature Encoding Via Spatio-Temporal Virtual Grid for 3D Object Detection
}
\author{Konyul Park$^{1}$$^{*}$, Yecheol Kim$^{2}$$^{*}$, Junho Koh$^{2}$, Byungwoo Park$^{1}$, and Jun Won Choi$^{3}$$^{\dagger}$
\thanks{$^{*}$These authors contributed equally to this work.}%
\thanks{$^{\dagger}$Corresponding author.}%
\thanks{$^{1}$Department of Artificial Intelligence, Hanyang University, Seoul,
04763, Korea.}%
\thanks{$^{2}$Department of Electrical Engineering, Hanyang University, Seoul,
04763, Korea.}
\thanks{$^{3}$Department of Electrical and Computer Engineering, College of Liberal Studies, Seoul National University, Seoul,
 08826, Korea.}
\thanks{{\tt\footnotesize \{konyulpark,yckim,jhkoh,bwpark\}@spa.hanyang.ac.kr}}
\thanks{{\tt\footnotesize junwchoi@snu.ac.kr}}
}
\begin{document}
\maketitle
\thispagestyle{empty}
\pagestyle{empty}

\begin{abstract}
Developing high-performance, real-time architectures for LiDAR-based 3D object detectors is essential for the successful commercialization of autonomous vehicles. Pillar-based methods stand out as a practical choice for onboard deployment due to their computational efficiency. However, despite their efficiency, these methods can sometimes underperform compared to alternative point encoding techniques such as Voxel-encoding or PointNet++. We argue that current pillar-based methods have not sufficiently captured the fine-grained distributions of LiDAR points within each pillar structure. Consequently, there exists considerable room for improvement in pillar feature encoding. In this paper, we introduce a novel pillar encoding architecture referred to as \textit{Fine-Grained Pillar Feature Encoding} (FG-PFE). FG-PFE utilizes \textit{Spatio-Temporal Virtual} (STV) grids to capture the distribution of point clouds within each pillar across vertical, temporal, and horizontal dimensions. Through STV grids, points within each pillar are individually encoded using Vertical PFE (V-PFE), Temporal PFE (T-PFE), and Horizontal PFE (H-PFE). These encoded features are then aggregated through an \textit{Attentive Pillar Aggregation} method. Our experiments conducted on the nuScenes dataset demonstrate that FG-PFE achieves significant performance improvements over baseline models such as PointPillar, CenterPoint-Pillar, and PillarNet, with only a minor increase in computational overhead.
\end{abstract}

\section{INTRODUCTION}
Ensuring the safety and stability of autonomous vehicles depends on their ability to perceive 3D spaces both accurately and in real-time. Recently, the advent of deep learning has spurred significant advancements in the field of LiDAR-based 3D object detection. While the majority of existing research has focused  on improving accuracy, the need for real-time responsiveness cannot be overstated in the real-world scenarios. Therefore, there is a growing need for research that prioritises the development of computationally efficient, real-time LiDAR-based 3D object detectors.

\begin{figure}
   
    {\includegraphics[width=85mm]{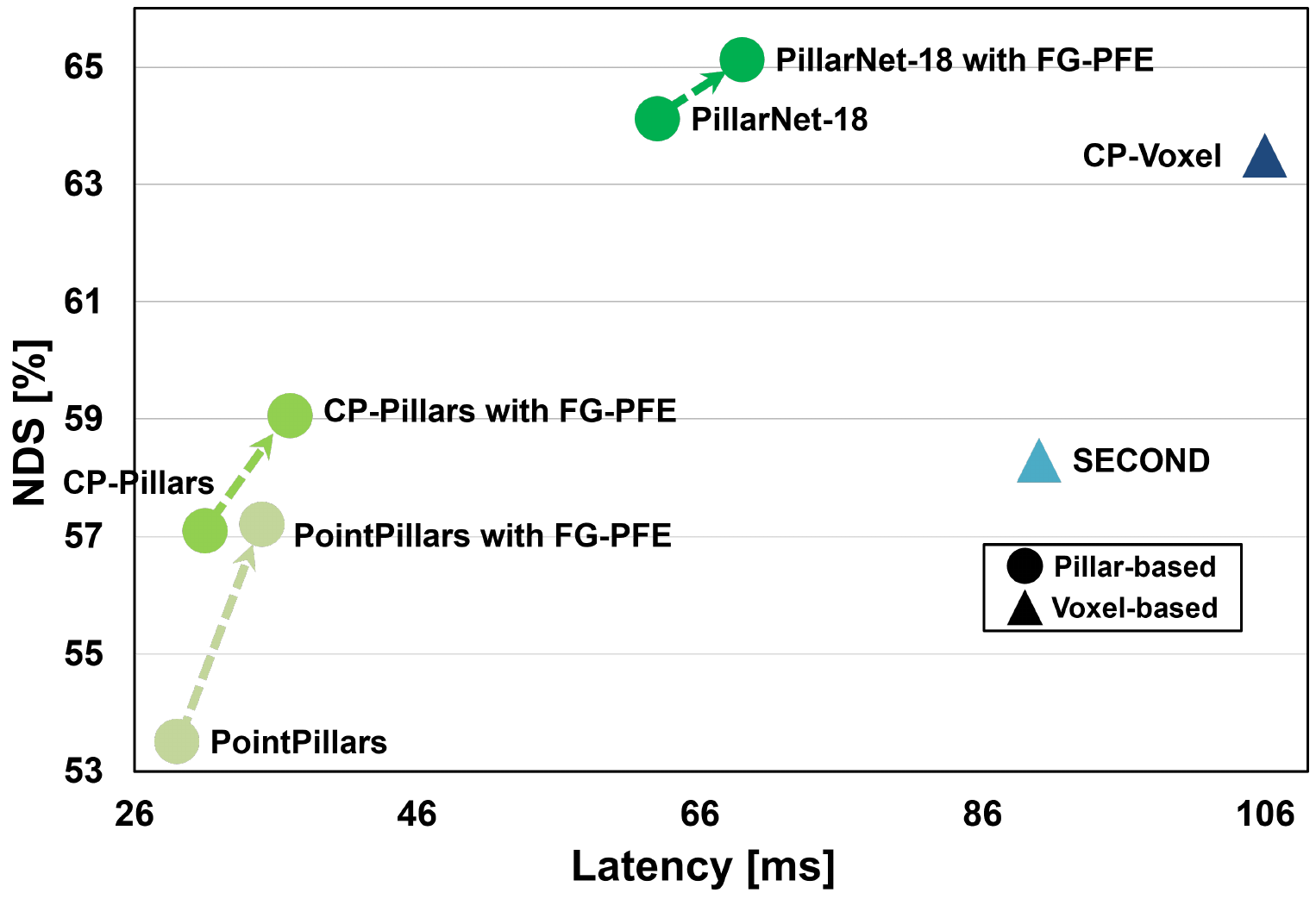}}
    \caption {\textbf{Performance versus latency of several 3D object detectors evaluated on nuScenes \textit{val} split}:  
    CP-Pillars denotes CenterPoint \cite{centerpoint} with PointPillars backbone. Latency is measured with a single NVIDIA TITAN RTX GPU. When incorporated into various baselines such as PointPillars \cite{pointpillars}, CenterPoint \cite{centerpoint}, and PillarNet \cite{pillarnet}, FG-PFE delivers substantial performance gains with small computational overhead.}
    \begin{comment}
    
    \end{comment}
    \label{intro_fig}
\end{figure}

To date, a variety of successful LiDAR-based 3D object detectors have been proposed,  including VoxelNet \cite{voxelnet}, SECOND \cite{second}, PointPillar \cite{pointpillars}, PV-RCNN \cite{pvrcnn}, and CenterPoint \cite{centerpoint}.  Given the irregular and sparse nature of LiDAR point clouds, various architectures have been proposed to model point clouds. These methods can be roughly categorized into point-based methods \cite{pointrcnn, 3dssd}, 3D voxel-based methods \cite{second, voxelnet, voxelnext}, and 2D pillar-based methods \cite{pointpillars, pillarnet, pillarnext}. Point-based methods \cite{pointrcnn, 3dssd} generate features directly from LiDAR points, modeling their spatial distribution in a hierarchical manner. However, these techniques require large computational overhead due to iterative point clustering step. Both voxel-based and pillar-based methods group LiDAR points using a predefined grid structure and encode the points within each grid, resulting in features organized in uniform grids. Voxel-based methods \cite{second, voxelnet, voxelnext} employ a 3D regular voxel structure to cluster LiDAR points and separately encode the points for each voxel element.  Then, 3D convolution layers are applied to further increase an abstraction level of the voxel features. However, voxel-based techniques need to encode the points in numerous voxels at the same time, and the 3D convolution itself is also computationally demanding. On the other hand, Pillar-based methods \cite{pointpillars, pillarnet, pillarnext} organize LiDAR points into a 2D grid in the x-y plane that forms a series of pillars. These methods significantly reduce the number of grid elements and the resulting features can be efficiently processed by 2D convolutions. As a result, pillar-based methods offer enhanced efficiency, enabling real-time processing at speeds exceeding 10 FPS (frames per second). Given these advantages, such methods are now considered as a promising architectural choice for commercial applications.

While pillar-based methods are known for their computational efficiency, they often fail to outperform alternative approaches. This performance gap may be attributed to the insufficient capability to capture spatial point distributions of the pillar encoding. To bridge this gap, several studies, including PillarNet \cite{pillarnet}, PillarNext \cite{pillarnext}, and HVNet \cite{hvnet}, have focused on enhancing the effectiveness of pillar feature encoding without compromising its computational speed. Efforts to address these challenges have concentrated on optimizing two critical aspects of pillar-based methods: the generation of pillar features and their subsequent encoding through convolutional layers.
PillarNet \cite{pillarnet} and PillarNeXt \cite{pillarnext} incorporated hierarchical deep pillar feature
extraction and multi-scale fusion neck modules to improve the convolutional encoding. HVNet \cite{hvnet} enhanced pillar representation by integrating an attention mechanism into multi-scale pillars. 
In this study, our objective is to improve the pillar encoding process in order to enhance the accuracy of 3D object detection.
We hypothesize that the subpar performance of pillar-based methods compared to 3D voxel-based methods stems from not effectively capturing the fine-grained distributions of LiDAR points within each pillar. This is due to all points within each pillar being compressed into a single feature, failing to capture the dynamic point distribution across both spatial and temporal domains. 

We propose a new pillar feature encoding method, referred to as \textit{Fine-grained Pillar Feature Encoding} (FG-PFE). Our observation is that the reduced performance of pillar-based encoding relative to 3D voxel-based encoding originates from the absence of vertical grids when encoding the points in each pillar.
Inspired by this observation, we devise \textit{Spatio-Temporal Virtual} (STV) grids to obtain a fine-grained pillar representation. STV grids group points based on their vertical grid, temporal scanning grid, and horizontal displacement grid. Within each pillar, points are individually processed by three distinct modules: Vertical PFE (V-PFE), Temporal PFE (T-PFE), and Horizontal PFE (H-PFE). 
V-PFE divides each pillar using a vertical virtual grid and encodes the points within each virtual cluster separately. It then performs a weighted aggregation of the feature vectors from different vertical grids through a channel attention module. Although this grid structure may seem akin to that used in voxel-based methods, V-PFE generates a single feature vector for each pillar, thus retraining a computational advantage over voxel encoding. T-PFE organizes points in each pillar based on the LiDAR's scanning order. It encodes points from each scan order separately, incorporating the ability to model their temporal distribution.  H-PFE introduces pillar grids with various horizontal grid offsets to capture multiple perspectives of Bird's Eye View (BEV) features. 
Finally, the proposed FG-PFE aggregates the features derived from these three encoding modules adaptively using an {\it Attentive Pillar Aggregation} (APA) module.

We evaluate the performance of FG-PFE on the widely used nuScenes benchmark \cite{nuscenes}. Fig. \ref{intro_fig} shows the performance versus complexity plot of FG-PFE in comparison with three pillar-based baseline methods, PointPillars \cite{pointpillars}, CenterPoint \cite{centerpoint} and PillarNet \cite{pillarnet}. FG-PFE offers significant performance gains over all baseline methods. In particular, FG-PFE achieves a 3.7\% increase in the nuScenes detection score (NDS) over PointPillars and a 1.0\% improvement over PillarNet, all with small additional computational increases.

The key contributions of the paper are summarized as follows.
\begin{itemize}

    \item We present FG-PFE, a novel pillar feature encoding method that promises performance improvement for LiDAR-based 3D object detection.
    \item We devise a virtual grid structure termed STV grids that can capture the fine-grained distribution of LiDAR points within a pillar. 
    STV grids provide added granularity to represent points across vertical, temporal, and horizontal dimensions.
    \item We introduce three distinct modules: V-PFE, T-PFE, and H-PFE. Each module generates pillar features using virtual grids in the vertical, temporal, and horizontal dimensions, respectively.  These features are aggregated to produce the final pillar features used for subsequent 2D convolutional encoding.  
    \item Our evaluation demonstrates that by integrating the proposed FG-PFE into pillar-based baselines, we achieve substantial performance enhancements with only a minor increase in computational overhead.
\end{itemize}

\begin{figure*}[t]
    
    \centering
    {\includegraphics[width=0.95\textwidth]{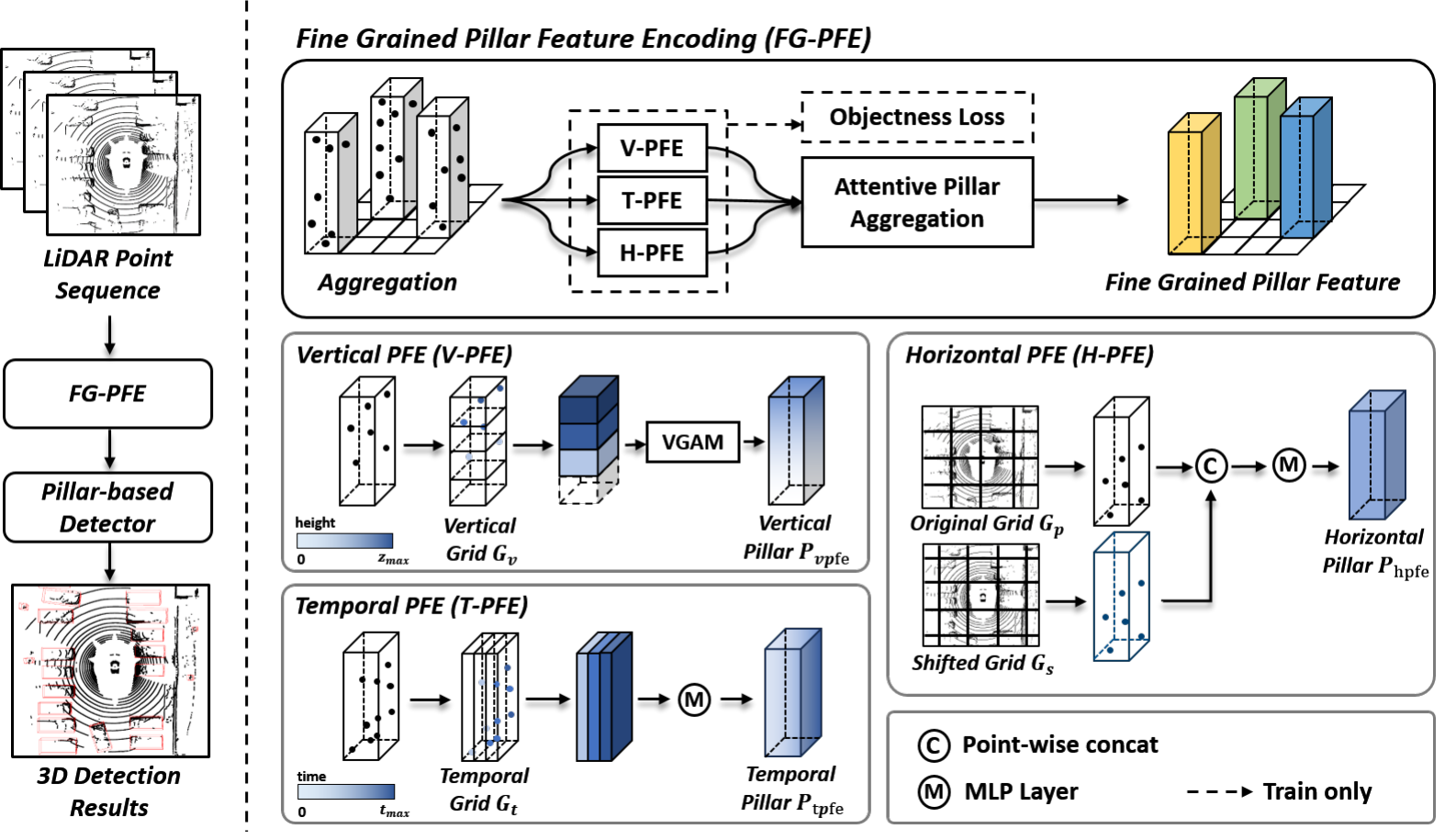 }}
    \hspace{5mm}
    \caption {\textbf{Overall architecture of the proposed FG-PFE.} LiDAR points are quantized along the vertical, temporal, and horizontal axes. In V-PFE, voxels from vertical axis are aggregated by the vertical grid attention module. In T-PFE, voxels from temporal axis are processed using a set of MLPs. In H-PFE, voxels from two different horizontal grid are transformed back into LiDAR points and then combined using concatenation. LiDAR points are then converted into pillar features with original pillar feature encoding. Pillar features originating from the three axes are combined using the Attentive Pillar Aggregation module.
    }
    
    \label{overall_fig}
\end{figure*}

\begin{comment}
\end{comment}

\section{RELATED WORK}
Numerous strategies have been developed for LiDAR-based 3D object detection, which generally fall into two main categories based on the type of input representation they use: point-based methods and grid-based methods.

Point-based techniques, such as PointRCNN \cite{pointrcnn} and  3DSSD \cite{3dssd},  directly process raw point clouds without requiring any preliminary processing. PointRCNN \cite{pointrcnn} utilized PointNet++ \cite{pointnet++} to first segment out the foreground 3D points and then refined the proposals using the segmented features. Meanwhile, 3DSSD \cite{3dssd} introduced an F-FPS method that complements the traditional D-FPS, incorporating a set abstraction operation that enhances both regression and classification tasks. Although point-based methods are known for their high spatial accuracy and the ability to flexibly aggregate features, their computational demands can be a limiting factor, especially in large-scale point cloud scenarios.

Grid-based approaches partition irregular point clouds using uniform grid and encode them individually for each grid element. Voxel-based methods, such as VoxelNet \cite{voxelnet} and SECOND \cite{second} employed 3D voxel structure to produce 4D voxel features, which were further encoded by 3D convolutional layers. Particularly, SECOND leveraged the inherent sparsity of voxel structures to reduce computational complexity.  Subsequent advancements, including FSD \cite{fsd} and VoxelNext \cite{voxelnext}, have significantly reduced computational demands by incorporating sparse features within the detection head. Despite these improvements, voxel-based methods still face challenges in real-time applications due to the need to process a large volume of voxels and the computational burden of 3D convolution operations. The pillar-based approach introduced in PointPillars \cite{pointpillars} organized point clouds into vertical columns called pillars. The features obtained from this pillar encoding were further processed through 2D convolution layers. Due to the absence of vertical grid and the use of 2D convolutions, this method significantly reduces computational load. 

Recent research efforts have been directed at addressing the modeling limitations of PointPillars. PillarNet \cite{pillarnet} and PillarNeXt \cite{pillarnext} have enhanced the performance of PointPillars by integrating 2D sparse convolution layers into the backbone encoder and incorporating multi-scale fusion neck modules. Furthermore, HVNet \cite{hvnet} and Voxel-FPN \cite{voxelfpn} have enhanced performance by combining pillar features from various scales at the point-wise level. SST \cite{sst} and DSVT \cite{dsvt} used shifted pillar grid and rotated-set pillar grids to allow information exchange across different encoding schemes. 

\section{PROPOSED METHOD}
In section, we present the details of the proposed FG-PFE.
\subsection{Fine-Grained Pillar Feature Encoding}
 The overall architecture of FG-PFE is depicted in Fig. \ref{overall_fig}. Initially, LiDAR point clouds are structured into a conventional pillar grid as described in PointPillars \cite{pointpillars}. Subsequently, the fine-grained distribution of points within each pillar is captured through simultaneous processing by V-PFE, T-PFE, and H-PFE. The features generated by these three encoding modules are then combined through the APA module. Consequently, the FG-PFE method generates pillar features that are of the same size as those produced by PointPillars. Finally, the consolidated pillar features are further processed by the subsequent stage of a typical 3D object detection framework.

% \noindent
\subsubsection{Vertical PFE} 
V-PFE organizes points within each pillar by employing a virtual vertical grid, then encodes them to generate the pillar feature $P_{vpfe}$.
Consider a scenario where the $i$th pillar contains a set of $N_i$ LiDAR points, denoted as $p\in\mathbb{R}^{N_i\times5}$. Each point is characterized by its $(x,y,z)$ coordinates, reflectance value, and time lag. For brevity, we omit the pillar index $i$ hereafter. As illustrated in Fig.~\ref{overall_fig}, a vertical grid structure $G_{v}$ with a size of $H_p$ partitions the pillar and the points within each vertical grid element are encoded using mean-pooling. The features in pillars that are not empty (i.e., containing more than one non-empty vertical grid element) are gathered, generating the features $P$ of dimensions $N_{vp}\times H_p \times C_p$, where $N_{vp}$ denotes the count of non-empty pillars, and $C_p$ denotes the channel dimension.
This feature volume $P$ is collapsed along a vertical dimension using the {\it Virtual Grid Aggregation Module} (VGAM). The overall VGAM process can be summarized as: 
\begin{gather}
P'=M_c(P) \otimes P, \\
P''=M_v(P') \otimes P', \\
P_{vpfe}=\psi(\text{reshape}(P'', N_{vp}, H_p \cdot C_p)),
\end{gather}
where \( \otimes \) denotes element-wise multiplication and $\psi ( \cdot )$ denotes MLP layer. The channel attention operation  $M_c(P)$ is expressed as
\begin{align}
M_c(P)&=\sigma(W_1(W_0(Avg(P))+W_1(W_0(Max(P))), \label{eqn:mc}
\end{align}
where \( \sigma \) is  the sigmoid function, $W_0\in\mathbb{R}^{{C/r}\times{C}}$, and $W_1\in\mathbb{R}^{{C}\times{C/r}}$, where $r$ is the reduction ratio. Note that MLP weights $W_0$ and $W_1$ are shared for both inputs and the RELU function is followed by $W_0$. The process of vertical attention is expressed as
\begin{align}
M_v(P)&=\sigma(f(concat(Avg(P),Max(P)))), \label{eqn:mv} 
\end{align}
where \( f(\cdot) \) represents a 1D convolutional layer with the filter size of 7. 
Note that   $P''$ is reshaped from the dimensions $N_{vp}\times H_p \times C_p$ to $N_{vp}\times (H_p \cdot C_p)$, and  an MLP layer is employed to process the reshaped features and restore the original channel dimensions $C_p$, yielding the vertical pillar feature $P_{vpfe}\in\mathbb{R}^{N_{vp}\times{C_p}}$.

% \noindent
\subsubsection{Temporal PFE}  T-PFE models the temporal distribution of LiDAR inputs obtained from multiple sweeps. The LiDAR points within each pillar are organized using a temporal grid structure $G_{t}$ of size $T_p$ such that each grid element contains points from different sweeps. Similar to V-PFE, the points within each grid element are encoded using mean-pooling.
The pillars containing more than one non-empty temporal grids are gathered, resulting in features of size $N_{tp}\times{T_p}\times{C_p}$, where $N_{tp}$ denotes the number of non-empty temporal pillars and $C_p$ denotes the channel dimension. % 
The tensor $P_{tp}$ is reshaped from $N_{tp}\times T_p \times C_{p}$ to $N_{tp}\times T_pC_p$ via grid-wise concatenation. Then, a MLP layer is used to transform back to its original channel dimension \( C_{p} \), resulting in \( P_{tpfe} \).

% \noindent
\subsubsection{Horizontal PFE}
The Horizontal PFE (H-PFE) utilizes multiple pillar grids to generate different pillar features. These grids are distinguished by different offsets in the horizontal domain. Consider two pillar grids shifted by half the grid size. Given a grid $G_p$, a shifted grid $G_s$ is derived by translating $G_p$ by half the grid size along the $x$ and $y$ axes. Subsequently, LiDAR points are partitioned using both the original and shifted grids, yielding the pillar features $P_{op}\in\mathbb{R}^{N_{p}\times{C_p}}$ and the shifted pillar features $P_{sp}\in\mathbb{R}^{N_{p}\times{C_p}}$. The pillar features obtained from the two distinct grids are concatenated in a point-wise manner. Finally, these combined features are arranged into the pillar grid $G_p$ to generate the horizontal pillar features $P_{hpfe}$.

% \noindent
\subsubsection{Attentive Pillar Aggregation}
The APA combines three distinct features obtained from V-PFE, T-PFE, and H-PFE. 
The APA initially reduces the channel dimension to facilitate efficient feature fusion. Subsequently, it employs channel-wise attention (CWA) \cite{senet} in conjunction with convolutional layers to selectively aggregate the three pillar features. The outcome is final pillar features $P_{fg}$ that maintain compatibility with existing pillar-based models. To summarize, the APA process unfolds as follows:
\begin{gather}
P_{fg}=\psi(CWA(concat(P_{vpfe},P_{tpfe},P_{hpfe}))),
\end{gather}
where $CWA(\cdot)$ denotes the channel-wise attention module.
\subsection{Objectness Score Prediction Loss}
\newcolumntype{C}{>{\centering\arraybackslash}p{2.1em}}
\newcolumntype{H}{>{\centering\arraybackslash}p{1.8em}}

\renewcommand{\arraystretch}{1.0}

\begin{table*}[t]
\begin{large}
\begin{center}
\caption{Quantitative comparison with state of the art methods On nuScenes test set. C.V and T.C presents the construction vehicle
and the traffic cone, respectively, PED. and MOTOR. are short for the pedestrian and the motorcycle, respectively. L indicates
the LiDAR. P/V/R denotes the pillar, voxel and range view based grid encoder, respectively. The best performance is bolded}
\begin{adjustbox}{width=1.0\linewidth}
{\normalsize
\begin{tabular}{c || c  | c  c  c |  c  c  c  c  c  c  c  c  c  c }
% \begin{tabular}{c || H  H |  C  C  C  C  C  C  C  C  C  C }
% \Xhline{4\arrayrulewidth}
\hline
Method & Encoder & mAP & NDS & Latency & Car & Truck
& Bus & Trailer & C.V & Ped. & Motor. & Bicycle & T.C & Barrier \\ \hline\hline

% SA-Det3D \cite{sa-det3d} & 59.2 & 47.0 & 81.2 & 43.8 & 57.2 & 47.8 & 11.3 & 73.3 & 32.1 & 7.9 & 60.6 & 55.3\\
CBGS \cite{cbgs} & V & 52.8 & 63.3 & - & 81.1 & 48.5 & 54.9 & 42.9 & 10.5 & 80.1 & 51.5 & 22.3 & 70.9 & 65.7\\
CVCNet \cite{cvcnet} & V+R & 55.8 & 64.2& - & 82.7 & 46.1 & 45.8 & 46.7 & 20.7 & 81.0 & 61.3 & 34.3 & 69.7 & 69.9\\
HotSpotNet \cite{hotspotnet} & V & 59.3 & 66.6& - & 83.1 & 50.9 & 56.4 & 53.3 & 23.0 & 81.3 & 63.5 & 36.6 & 73.0 & 71.6\\
CenterPoint \cite{centerpoint} & V & 58.0 & 65.5 & 106ms & 84.6 & 51.0 & 60.2 & 53.2 & 17.5 & 83.4 & 53.7 & 28.7 & 76.7 & 70.9\\
Focals Conv \cite{focalsconv} & V & 63.8 & 70.0 & - &86.7 &56.3 &67.7 &59.5 &23.8 &87.5 &64.5 &36.3 &81.4 &74.1 \\
AFDetV2 \cite{afdetv2} & V  & 62.4 & 68.5& -  &86.3 & 54.2 & 62.5 & 58.9 & 26.7 & 85.8 & 63.8 & 34.3 & 80.1 & 71.0 \\
UTVR-L \cite{uvtr} & V & 63.9 & 69.7&  & 86.3 & 52.2&62.8&59.7&33.7&84.5&68.8&41.1&74.7&74.9\\
VISTA \cite{vista} & V+R & 63.0 & 69.8  & - & 84.4 & 55.1 & 63.7 & 54.2 & 25.1 & 82.8 & 70.0 & 45.4 & 78.5 & 71.4\\
Transfusion-L \cite{transfusion} & V & 65.5 & 70.2  & - & 86.2 & 56.7 & 66.3 & 58.8 & 28.2 & 86.1 & 68.3 & 44.2 & 82.0 & 78.2\\
PointPillars \cite{pointpillars} & P & 30.5 & 45.3 & 31ms & 68.4 & 23.0 & 28.2 & 23.4 & 4.1 & 59.7 & 27.4 & 1.1 & 30.8 & 38.9\\
PillarNet-18 \cite{pillarnet} & P & 65.0 &70.8  & 63ms & 87.4 &56.7 &60.9 &61.8 &30.4 &87.2 &67.4 &40.3 &82.1 &76.0\\
%VoxelNext \cite{voxelnext} & V & 66.2 &71.4  & - & 85.3 &55.7 &66.2 &57.2 &29.8 &86.5 &75.2 &48.8 &80.7 &76.1\\
\hline %\hline

Ours & P & \textbf{65.7} & \textbf{71.8} & 69ms & 85.2 & 55.5 & 61.6 & 61.8 & 29.2 & 86.5 & 72.5 & 48.6 & 82.5 & 73.8\\

\hline
% \Xhline{4\arrayrulewidth}

\end{tabular}
\normalsize}
\end{adjustbox}
\label{table:main_result}
\end{center}
\end{large}
\end{table*}
\renewcommand{\arraystretch}{1}

To enhance the individual features generated from the virtual grids, we introduce the concept of \textit{objectness score prediction loss}, which compels the model to predict objectness scores for each virtual grid element. This loss encourages the model to discriminate between foreground and background grids, thus improving scene representation. The segmentation labels $y_{seg}$ are directly derived from the 3D detection box annotations, indicating whether each grid is within or outside a ground-truth 3D box.   The objectness score prediction is conducted by passing each pillar feature through a single MLP layer followed by a sigmoid function. The objectness score prediction loss function is given by 
\begin{align}
  \mathcal{L}_{op}=\lambda_{gl}(l_{op}(P_{vpfe})+l_{op}(P_{tpfe})+l_{op}(P_{hpfe})), 
  \end{align}
  where
\begin{align}
 l_{op}(P) = FL(\mathcal{A}(P)), y_{seg}), 
 \end{align}
and $FL(\cdot)$ denotes the focal loss function \cite{focal}, $\mathcal{A}(\cdot)$ represents an MLP layer followed by a sigmoid function used for predicting objectness scores, and $\lambda_{gl}$ is the balanced loss factor. The total loss for the entire network is the sum of the detection loss $\mathcal{L}_{det}$ and the objectness score prediction loss $\mathcal{L}_{op}$.

\subsection{Modification on Group Head}
The group head proposed in CBGS \cite{cbgs} organizes classes into hierarchical groups based on similar shapes and sizes. This allows the model distinguish objects within groups that share similar shape and size attributes. However, our experiments found that when we use separate group heads for bicycle and motorcycle classes, we observed a significant improvement in Average Precision (AP) over conventional grouping. This suggests that grouping classes of similar attributes like motorcycles and bicycles together does not necessarily lead to better performance. Our modified group head shows notable performance gains compared to conventional designs.

\section{EXPERIMENTS}
\subsection{nuScenes Dataset}
% \noindent
The nuScenes dataset \cite{nuscenes} constitutes a large-scale autonomous driving dataset comprising 1000 scenes, which consist of 700 training scenes, 150 validation scenes, and 150 test scenes. Each scene comprises LiDAR point cloud data acquired at a rate of $20 Hz$ using a 32-channel LiDAR system. Annotated samples are provided at $2 Hz$ and contain 1.4 million 3D bounding boxes. We follow official evaluation protocol for 3D object detection and evaluate mean Average Precision (mAP)
and the nuScenes Detection score (NDS) across 10 foreground classes: barrier, bicycle, bus, car, motorcycle, pedestrian, trailer, truck, construction vehicle, and traffic cone.

\subsection{Implementation Details}
We integrate our FG-PFE module with PillarNet-18 \cite{pillarnet}, which is a state-of-the-art pillar-based detector. 
Each LiDAR point is represented as a 5-dimensional vector, $(x, y, z, r, \Delta t)$, where $(x, y, z)$ denotes the 3D coordinate of the point, $r$ indicates the point's intensity, and $\Delta t$ represents the temporal displacement from the keyframe within the range $[0, 0.5)$. We consider a detection range spanning $[-54m, 54m]$ along the X and Y axes, and $[-5m, 3m]$ along the Z axis. The LiDAR pillar grid maintains a spatial resolution of $(0.075m, 0.075m)$, resulting in a pillar structure composed of $1440 \times 1440$ pillars.
\newcolumntype{E}{>{\centering\arraybackslash}p{6em}}
\renewcommand{\arraystretch}{1.3}

\begin{table}[t]
\begin{center}
\caption{The effects of FG-PFE modules: STV, objectness score prediction loss and Modified group head.}
\begin{adjustbox}{width=0.49\textwidth}
\resizebox{\columnwidth}{!}{%
\begin{tabular}{c||ccc||c|c}
\hline
% \Xhline{4\arrayrulewidth}
\multirow{2}{*}{\it { Method}}      & \multicolumn{3}{c||}{Module} & \multicolumn{2}{c}{Performance}\\  \cline{2-6} 
                                    & {STV}                     & {Loss}      & {Head}     & {NDS (\%)}               & {Latency}               \\ \hline \hline
PillarNet-18                        &                              &             &            & {64.12}                  & {63ms}                  \\ \hline
\multirow{3}{*}{{\small Ours}} & {\checkmark}                 &             &            & {65.13$_{\uparrow 1.01}$}& {68ms}\\ 
                                    & {\checkmark}                 & {\checkmark}&            &  65.43$_{\uparrow 1.31}$  & {68ms}  \\ 
                                    & {\checkmark}                 & {\checkmark}&{\checkmark}& 65.58$_{\uparrow 1.46}$  & {69ms}  \\ \hline
%SECOND                     &        &      &   & {-} & {-}             \\ \hline

% \Xhline{4\arrayrulewidth}
% \hline
\end{tabular}%
}

\end{adjustbox}
% \caption{{\bf Ablation study to evaluate the sub-modules of SM-VFE.}}
\label{table:main_ablation}
\end{center}

\end{table}

\renewcommand{\arraystretch}{1}
\newcolumntype{E}{>{\centering\arraybackslash}p{6em}}
\renewcommand{\arraystretch}{1.3}

\begin{table}[t]
\begin{center}
\caption{The effects of proposed V-PFE, T-PFE and H-PFE in STV.}
\begin{adjustbox}{width=0.49\textwidth}
\resizebox{\columnwidth}{!}{%
\begin{tabular}{c||ccc||c|c}
\hline
%\Xhline{4\arrayrulewidth}
\multirow{2}{*}{{\it Method}}    & \multicolumn{3}{c||}{Module} & \multicolumn{2}{c}{Performance} \\ \cline{2-6} 
                                 & {V-PFE}      &{T-PFE}      & {H-PFE}           & {NDS (\%)}               & {Latency}              \\ \hline \hline   
PillarNet-18                     &              &             &                   & {64.12}                  & {63ms}                 \\ \hline
\multirow{3}{*}{{\small FG-PFE}}          & {\checkmark} &             &          & {64.50$_{\uparrow 0.38}$}& {65ms}  \\ 
                                 & {\checkmark} & {\checkmark}&                   & 64.83$_{\uparrow 0.71}$  & {66ms}  \\
                                 & {\checkmark} & {\checkmark}&{\checkmark}       & 65.13$_{\uparrow 1.01}$  & {68ms}  \\   \hline 
%SECOND                     &        &      &    & {-} & {-}    \\ \hline
% \Xhline{4\arrayrulewidth}
% \hline
\end{tabular}%
}

\end{adjustbox}
% \caption{{\bf Ablation study to evaluate the sub-modules of SM-VFE.}}
\label{table:pfe_ablation}
\end{center}
\end{table}

\renewcommand{\arraystretch}{1}

We trained our model from scratch using a batch size of 16. For optimization, we adopted a one-cycle learning rate policy spanning 20 epochs, with a maximum learning rate set to 0.001. Furthermore, we employed a data augmentation strategy encompassing random flipping, rotation, scaling, and ground truth box sampling \cite{second}. We set the balanced loss factor  $\lambda_{gl}$ to 1. In the inference stage, we optimize performance by employing batch normalization folding.  During evaluation on the nuScenes test set, we utilized the double-flip test-time augmentation technique, following the method proposed in CenterPoint. \cite{centerpoint}. The model's latency was measured on a single NVIDIA TITAN RTX GPU. 

\subsection{Performance on nuScenes Test Set}
Table \ref{table:main_result} presents a comparative analysis of our model against previous LiDAR-only 3D object detectors on the nuScenes test set. The performance metrics of other LiDAR-based methods are sourced from the nuScenes leaderboard. Our model demonstrates superior performance, achieving an mAP of 65.7\% and an NDS of 71.8\%, surpassing other LiDAR-based approaches. Notably, our model exhibits an improvement of 0.7\% in mAP and 1.0\% in NDS over the PillarNet-18 baseline, while introducing a negligible runtime increase of only 6 milliseconds. This result highlights a significant enhancement in performance, particularly in the detection of bicycles, motorcycles, traffic cones, and buses.

\subsection{Ablation Studies on nuScenes Valid Set}
In this section, we conduct several ablation studies on the nuScenes validation set. To expedite these experiments, we trained our models using only 25\% of the training set, while utilizing the entire validation set for evaluation.

\noindent
\subsubsection{Contribution of Main Modules}
Table \ref{table:main_ablation} illustrates the impact of STV, objectness score prediction loss, and the modified group head on overall performance. Integrating spatio-temporal information in the pillar encoding stage through STV yields a notable 1.01\% enhancement in NDS with a mere $5ms$ latency. Furthermore,  the objectness score prediction loss contributes an additional 0.3\% improvement in NDS without incurring extra computation. Lastly, the incorporation of the modified group head to discriminate finer details results in a 0.15\% increase in NDS with a latency of $2ms$. The combination of all three components yields a cumulative performance gain of 1.46\% in NDS over the PillarNet baseline. This underscores the significant enhancement in detection performance achieved by the proposed method while maintaining computational efficiency

\noindent
\subsubsection{Contribution of V-PFE, T-PFE and H-PFE}
Table \ref{table:pfe_ablation} shows the contribution of each pillar encoding module within the FG-PFE to the overall performance. The V-PFE module, which is designed to encode the point distribution in vertical dimension within each pillar, provides an increase of 0.38\% in NDS. The integration of T-PFE with V-PFE improves the NDS by 0.71\%. Finally, the integration of all three modules boils down to a total performance gain of 1.01\% in NDS over the PillarNet baseline.

\noindent
\subsubsection{Performance Gain over Different Baselines}
Table \ref{table:ablation_pillarmodel} presents the 3D object detection performance when FG-PFE is integrated into several pillar-based 3D detection baselines. Integrating FG-PFE into the PillarNet-18 model yields improvements of 1.06\% in mAP and 1.01\% in NDS. Moreover, when applied to the pillar-based CenterPoint model, FG-PFE provides higher performance gains of 2.65\% in mAP and 1.96\% in NDS. The integration of FG-PFE with the PointPillars method demonstrates a notable improvement of 4.12\% in mAP and 3.7\% in NDS. The increase in latency due to FG-PFE is negligible, ranging from only $5ms$ to $6ms$, underscoring the efficiency of the method across various 3D detection methods. These results conclusively demonstrate that FG-PFE consistently delivers performance enhancements across a spectrum of baseline models.

\newcolumntype{E}{>{\centering\arraybackslash}p{10.6em}}
% \newcolumntype{F}{>{\centering\arraybackslash}p{4.5em}}
\renewcommand{\arraystretch}{1.3}

\begin{table}[t]
\begin{center}
\caption{The gains of FG-PFE on several baselines.}
\begin{adjustbox}{width=0.49\textwidth}
\begin{tabular}{E||c|c|c}
\hline
% \Xhline{4\arrayrulewidth}  {51.22$_{\uparrow 1.89}$}
\multirow{2}{*}{{\it Method}} & \multicolumn{3}{c}{Performance}\\ \cline{2-4}
&  mAP (\%) & NDS (\%) & Latency \\ \hline \hline

PillarNet-18  & 55.26 & 64.12 & 63ms \\ \hline

+ FG-PFE  & 56.32$_{\uparrow 1.06}$  & 65.13$_{\uparrow 1.01}$  & 68ms \\ \hline

CenterPoint-PointPillars & 45.30 & 57.10 & 31ms \\ \hline

+ FG-PFE & 47.95$_{\uparrow 2.65}$ & 59.06$_{\uparrow 1.96}$ & 36ms \\ \hline

PointPillars & 39.47 & 53.51 & 29ms \\ \hline

+ FG-PFE & 43.59$_{\uparrow 4.12}$ & 57.21$_{\uparrow 3.7}$ & 35ms \\ \hline

% \Xhline{4\arrayrulewidth}https://www.overleaf.com/project/64c109e5b47c08d941a4a05b

\end{tabular}
\end{adjustbox}
% \caption{{\bf Ablation study to evaluate the sub-modules of SM-VFE.}}
\label{table:ablation_pillarmodel}
\end{center}
\end{table}

\renewcommand{\arraystretch}{1}

\section{CONCLUSIONS}
In this paper, we introduced FG-PFE, a novel approach to pillar feature encoding designed to effectively capture the dynamic distribution of point clouds across spatial and temporal dimensions. Our method leverages STV grids to achieve a fine-grained pillar representation of point cloud distributions in vertical, temporal, and horizontal dimensions. V-PFE enhances the point cloud representation by segmenting each pillar further using a vertical virtual grid.  T-PFE encodes points obtained from different scanning orders, capturing the temporal distribution of the point clouds.  H-PFE explores multiple perspectives in representing the BEV scenes by employing different horizontal grid offsets. We introduced an objectness score prediction loss that aligns the three modules towards a unified detection objective. As a plug-and-play approach, FG-PFE can be readily integrated into existing pillar-based detectors. Our evaluation on the nuScenes dataset demonstrates that FG-PFE delivers significant improvements over several pillar-based baselines with only a negligible increase in computational complexity.

\section{ACKNOWLEDGMENT}

This work was supported by National Research Foundation of Korea (NRF) grant funded by the Korea government (MSIT) (No.2020R1A2C2012146).

\addtolength{\textheight}{0cm}   

\bibliographystyle{IEEEtran}
%\bibliography{LaTeX/reference}

\end{document}